\documentclass[10pt,twocolumn,letterpaper]{article}

\makeatother
\usepackage{wacv}
\usepackage{times}
\usepackage{wrapfig}
\usepackage{color}
\usepackage{times}
\usepackage{epsfig}
\usepackage{epstopdf}
\usepackage{graphicx}
\usepackage{amsmath}
\usepackage{amssymb}
\usepackage{multirow}
\DeclareGraphicsExtensions{.eps}

\usepackage[font=small,labelfont=bf]{caption}
\usepackage{subcaption}
\usepackage[colorlinks=true,bookmarks=false]{hyperref}
\allowdisplaybreaks
\wacvfinalcopy 
\def\wacvPaperID{283} 

\ifwacvfinal\fi
\begin{document}

\title{Unifying Registration based Tracking: A Case Study with Structural Similarity}

\author{Abhineet Singh \hspace{2cm} Mennatullah Siam \hspace{2cm} Martin Jagersand \\
University of Alberta\\
{\tt\small asingh1,mennatul,mj7@ualberta.ca}
}

\maketitle
\ifwacvfinal\thispagestyle{empty}\fi



\begin{abstract}
This paper adapts a popular image quality measure called structural similarity for high precision registration based tracking while also introducing a simpler and faster variant of the same. Further, these are evaluated comprehensively against existing measures using a unified approach to study registration based trackers that decomposes them into three constituent sub modules - appearance model, state space model and search method.
Several popular trackers in literature are broken down using this method so that their contributions - as of this paper - are shown to be limited to only one or two of these submodules.



An open source tracking framework  is made available that follows this decomposition closely through extensive use of generic programming. It is used to perform all experiments on four publicly available datasets so the results are easily reproducible.
This framework provides a convenient interface to plug in a new method for any sub module and combine it with existing methods for the other two. It can also serve as a fast and flexible solution for practical tracking needs due to its highly efficient implementation.
\end{abstract}

\section{Introduction}
\label{introduction}
Visual tracking is an important field in computer vision with diverse application domains including robotics, surveillance, targeting systems, autonomous navigation and augmented reality. Registration based tracking (\textbf{RBT}),
also known in literature as direct visual tracking \cite{Silveira10_esm_ext2, Scandaroli2012_ncc_tracking, Richa12_robust_similarity_measures, Richa14_scv_constraints},
is a
sub field thereof where the object is tracked by warping each image in a sequence to align the object patch with the template. Trackers of this type are especially popular in robotics and augmented reality applications both because they estimate the object pose with greater precision and are significantly faster than online learning and detection based trackers (\textbf{OLT}s) \cite{singh16_modular_results}.

However, OLTs are better suited to general purpose tracking as RBTs are prone to failure when the object undergoes significant appearance changes due to factors like occlusions and  lighting variations or when completely novel views of the object are presented by deformations or large pose changes. 
As a result, OLTs are more popular in literature and have been the subject of several recent studies \cite{Liu14_olt_survey, Smeulders2014_tracking_survey, Kristan2016_vot16}. The scope of such studies is usually limited to finding the trackers that work best under
challenging conditions by testing them on representative sequences with little to no analysis provided regarding \textit{why} specific trackers work better than others for given challenges. This is understandable since such trackers differ widely in design and have little in common that may be used to relate them to each other and perform comparative analysis from a design perspective.

As we show in this work, however, RBTs do not have this drawback and can be decomposed into three sub modules - appearance model (\textbf{AM}), state space model (\textbf{SSM}) and search method (\textbf{SM}) (Sec. \ref{decompose_registration_tracking}) - that makes their systematic analysis feasible. Though this decomposition is somewhat obvious and indeed has been observed before \cite{Szeliski2006_fclk_extended, Richa12_robust_similarity_measures}, it has never been explored systematically or used to improve the study of this paradigm of tracking. It is the intent of this work to fill this gap by unifying the myriad of contributions made in this domain since the original Lucas Kanade (LK) tracker was introduced \cite{Lucas81lucasKanade}.
Most of these works have presented novel ideas for only one of these submodules while using existing methods, often selected arbitrarily, for the rest.
For instance, Hager \& Belheumer \cite{Hager98parametricModels}, Shum \& Szeliski\cite{Shum00_fc}, Baker \& Matthews\cite{Baker01ict} and Benhimane \& Malis\cite{Benhimane04esmTracking} all introduced new variants of the Newton type SM used in \cite{Lucas81lucasKanade} but only tested these with SSD\footnote{refer Sec. \ref{decompose_registration_tracking} for acronyms} AM.
Similarly, Dick et. al \cite{Dick13nn} and Zhang et. al \cite{Zhang2015_rklt} only combined their stochastic SMs with RSCV and SSD respectively.
Conversely, Richa et. al \cite{Richa11_scv_original}, Scandaroli et. al \cite{Scandaroli2012_ncc_tracking} and Dame et. al \cite{Dame10_mi_ict} introduced SCV, MI and NCC as AMs but combined these only with a single SM - ESM in the former and ICLK in the other two.
Even more recent works that use illumination models (\textbf{ILM}) (Sec. \ref{appearanceModel}), including Bartoli \cite{Bartoli2006,Bartoli2008}, Silvera \& Malis \cite{Silveira2007_esm_lighting,Silveira2009_ilm_rgb_vs,Silveira2010_esm_ilm_rgb} and Silvera \cite{Silveira14_esm_ilm_omni}, have combined their respective ILMs with only a single SM in each case.
Finally, most SMs and AMs have been tested with only one SSM - either homography \cite{Benhimane04esmTracking,Benhimane07_esm_journal,Dame10_mi_ict} or affine \cite{Baker01ict,Ross08ivt}.
In fact, Benhimane \& Malis \cite{Benhimane04esmTracking} mentioned that their SM only works with  $\mathbb{SL}(3)$ SSM though experiments (Sec. \ref{results}) showed that it works equally well with others.


Such limited testing might not only give false indications about a method's capability but might also prevent it from achieving its full potential. For instance, an AM that outperforms others with a given SM might not do so with other SMs and an SM may perform better with an AM other than the one it is tested with. In such cases, our decomposition can be used to experimentally find the optimal combination of methods for any contribution while also providing novel insights about its workings. We demonstrate its practical applicability by comparing several existing methods for these sub modules not only with each other but also with two new AMs based on structural similarity (\textbf{SSIM}) \cite{Wang04_ssim_original} that we introduce here and fit within this framework.

SSIM is a popular measure for evaluating the quality of image compression algorithms by comparing the compressed image with the original one.
Since it measures the information loss in the former - essentially a slightly distorted or damaged version of the latter - it makes a suitable metric for comparing candidate warped patches with the template to find the one with the minimum loss and thus most likely to represent the same object. Further, it has been designed to capture the perceptual similarity of images and is known to be robust to illumination and contrast changes \cite{Wang04_ssim_original}. It is reasonable, therefore, to expect it to perform well for tracking under challenging conditions too. As such, it has indeed been used for tracking before with particle filters \cite{Loza2009_ssim_pf,Wang2013_ssim_bc}, gradient ascent \cite{Loza11_ssim_lk_tracking} and hybrid \cite{Loza2009a_ssim_hybrid} SMs. All of these trackers, however, used imprecise SSMs  with low degrees of freedom (DOF) - estimating only translation and scaling of the target patch. To the best of our knowledge, no attempt has been made to use SSIM for high DOF RBT within the LK framework \cite{Lucas81lucasKanade}. This work aims to fill this gap too.

To summarize, following are the main contributions of this work:
\begin{itemize}
\item Adapt a popular image quality measure - SSIM - for high precision RBT and introduce a simpler but faster version called SPSS (Sec. \ref{sec_ssim}).

\item Evaluate these models comprehensively by comparing against 8 existing AMs using 11 SMs and 7 SSMs. Experiments are done using 4 large datasets with over 100,000 frames in all to ensure their statistical significance.

\item Compare low DOF RBTs against state of the art OLTs to validate the suitability of the former for fast and high precision tracking applications.

\item Provide an open source tracking framework\footnote{available at \url{http://webdocs.cs.ualberta.ca/~vis/mtf/}}
called MTF \cite{singh16_mtf} using which all results can be reproduced and which, owing to its efficient C++ implementation, can also serve as a practical tracking solution.
\end{itemize}
Rest of this paper is organized as follows - section \ref{background} describes the decomposition, section \ref{sec_ssim} presents details about SSIM and section \ref{experiments} provides results and analysis.
\section{Background}
\label{background}
\subsection{Notation}
Let $ I_t : \mathbb R^2\mapsto \mathbb R $ refer to an image captured at time $t$ treated as a function of real values using sub pixel interpolation \cite{Dame10_mi_thesis} for non integral locations.
The patch corresponding to the tracked object's location in $ I_t $ is denoted by  $\mathbf{I_t}(\mathbf{x_t}) \in \mathbb R^N $ where $\mathbf{x_t}=[\mathbf{x_{1t}},..., \mathbf{x_{Nt}}]$ with $\mathbf{x_{kt}}=[x_{kt}, y_{kt}]^T \in  \mathbb R^2$ being the Cartesian coordinates of pixel $ k $.
Further, $\mathbf{w}(\mathbf{x}, \mathbf{p_s}) : \mathbb{R}^2 \times \mathbb{R}^S\mapsto \mathbb{R}^2$ denotes a warping function of $ S $ parameters
that represents the set of allowable image motions of the tracked object by specifying the deformations that can be applied to $\mathbf{x_0}$ to align
$\mathbf{I_t}(\mathbf{w}(\mathbf{x_0},\mathbf{p_{st}}))$
with
$ \mathbf{I_0}(\mathbf{x_0}) $.
Finally $f(\mathbf{I^*}, \mathbf{I^c},\mathbf{p_a} ) : \mathbb{R}^N \times \mathbb{R}^N \times \mathbb{R}^A\mapsto \mathbb{R}$ is a function of $A$ parameters that measures the similarity between two
patches - the reference or template patch $\mathbf{I^*}$
and a candidate patch $ \mathbf{I^c} $.
For brevity, $ \mathbf{I_t}(\mathbf{w}(\mathbf{x_0}, \mathbf{p_{st}}))$ and $\mathbf{I_0}(\mathbf{x_0})$ will also  be denoted as $ \mathbf{I_t} $ and $ \mathbf{I_0} $ respectively.

\subsection{Decomposing registration based tracking}
\label{decompose_registration_tracking}
Using this notation, RBT can be formulated as a search problem whose aim is to find the optimal parameters $\mathbf{p_t}=[\mathbf{p_{st}}, \mathbf{p_{at}}]\in\mathbb{R}^{S+A}$
that maximize the similarity, given by $f$, between $\mathbf{I^*} = \mathbf{I_0}$
and $\mathbf{I^c}=\mathbf{I_t}$, that is,
\begin{equation}
\begin{aligned}
\label{reg_tracking}
\mathbf{p_t} = \underset{\mathbf{p_s},\mathbf{p_a} } {\mathrm{argmax}} ~f(\mathbf{I_0}(\mathbf{x_0}),\mathbf{I_t}(\mathbf{w}(\mathbf{x_0},\mathbf{p_s})), \mathbf{p_a})
\end{aligned}
\end{equation}
This formulation gives rise to an intuitive way to decompose the tracking task into three sub modules - the similarity metric $ f $, the warping function $ \mathbf{w} $ and  the optimization approach. These can be designed to be  semi independent in the sense that any given optimizer can be applied unchanged to several combinations of methods for the other two modules which in turn interact only through a well defined and consistent interface. 
In this work, these sub modules are respectively referred to as AM, SSM and SM. A more detailed description with examples follows.

\subsubsection{Appearance Model}
\label{appearanceModel}

This is the image similarity metric defined by $f$ in Eq. \ref{reg_tracking} that the SM uses to compare different warped patches from $ I_t $ to find the closest match with  $\mathbf{I^*}$. 
It may be noted that $\mathbf{I^*}$ is not constrained to be fixed but may be updated during tracking as in \cite{Ross08ivt}.
Examples of $f$ with $A=0$ include sum of squared differences (\textbf{SSD}) \cite{Lucas81lucasKanade,Baker04lucasKanade_paper,Benhimane07_esm_journal}, sum of conditional variance (\textbf{SCV}) \cite{Richa11_scv_original}, reversed SCV (\textbf{RSCV}) \cite{Dick13nn}, normalized cross correlation (\textbf{NCC}) \cite{Scandaroli2012_ncc_tracking}, mutual information (\textbf{MI}) \cite{Dowson08_mi_ict,Dame10_mi_ict} and cross cumulative residual entropy (\textbf{CCRE}) \cite{Wang2007_ccre_registration,Richa12_robust_similarity_measures}.
There has also been a recent extension to SCV called \textbf{LSCV} \cite{Richa14_scv_constraints} that claims to handle localized intensity changes better.
Finally, it has been shown \cite{Ruthotto2010_thes_ncc_equivalence} that SSD performs better when applied to z-score \cite{Jain2005_ncc_z_score} normalized images which makes it equivalent to NCC. The resultant formulation is considered here as a distinct AM called Zero Mean NCC (\textbf{ZNCC}).
These AMs can be divided into 2 categories - those that use some form of the L2 norm as $ f $ - SSD, SCV, RSCV, LSCV and ZNCC - and those that do not - MI, CCRE and NCC. The latter are henceforth called robust AMs after \cite{Richa12_robust_similarity_measures}. The two AMs introduced here - SSIM and SPSS - too fall into this category.

To the best of our knowledge, the only AMs with  $A\neq 0$ introduced thus far in literature are those with an illumination model (\textbf{ILM}), where $f$ is expressed as $f(\mathbf{I^*}, g(\mathbf{I^c},\mathbf{p_a}))$ with $g : \mathbb{R}^N \times \mathbb{R}^A \mapsto \mathbb{R}^N$ accounting for differences in lighting conditions under which $I_0$ and $I_t$ were captured. These include gain \& bias (\textbf{GB}) \cite{Bartoli2006,Bartoli2008} and piecewise gain \& bias (\textbf{PGB}) \cite{Silveira2007_esm_lighting,Silveira2009_ilm_rgb_vs,Silveira2010_esm_ilm_rgb,Silveira14_esm_ilm_omni} with the latter comprising surface modeling with radial basis function (\textbf{RBF}) too.  Though there have also been AMs that incorporate online learning \cite{Jepson2003_online_am,Ross08ivt,Firouzi2014227_online_am}, they are not compatible with all SMs used here and so have not been tested.


\subsubsection{State Space Model}
\label{stateSpace}
This is the warping function $ \mathbf{w} $ that represents the deformations that the tracked patch can undergo. It therefore defines the set of allowable image motions of the corresponding object and can be used to place constraints on the search space of $ \mathbf{p_s} $ to make the optimization more robust or efficient. Besides the DOF of allowed motion, SSM also includes the actual parameterization of $ \mathbf{w} $. For instance, even though both represent 8 DOF motion, $\mathbb{SL}(3)$ \cite{Benhimane07_esm_journal, Kwon2014_sl3_aff_pf} is considered as a different SSM than standard homography \cite{Lucas81lucasKanade,Baker04lucasKanade_paper} that uses actual entries of the corresponding matrix.

This work uses 7 SSMs including 5 from the standard hierarchy of geometrical transformations \cite{Hartley04MVGCV,Szeliski2006_fclk_extended} - translation, isometry, similitude, affine and homography - along with two alternative parameterizations of homography - $\mathbb{SL}(3)$ and corner based (i.e. using x,y coordinates of the bounding box corners). More complex SSMs  to handle non rigid objects have also been proposed in literature like thin plate splines \cite{Bookstein1989_tps},  basis splines \cite{Szeliski1997_spline} and quadric surfaces \cite{Shashua2001_q_warp}. However, these are not tested here as the datasets used only feature rigid objects so these cannot be fairly evaluated.
Extensions like incorporation of 3D pose \cite{Cobzas2009_ic_3d_tracking} and camera parameters \cite{Trummer2008_gklt} to $ \mathbf{w} $ are also excluded.



\subsubsection{Search Method}
\label{searchMethod}
This is the optimization procedure that searches for the warped patch in $ I_t $ that best matches $\mathbf{I^*}$. There have been two main categories of SMs in literature - gradient descent (GD) and stochastic search. The former category includes the four variants of the classic Lucas Kanade (LK) tracker \cite{Lucas81lucasKanade} - forward additive (\textbf{FALK}) \cite{Lucas81lucasKanade}, inverse additive (\textbf{IALK}) \cite{Hager98parametricModels},  forward compositional (\textbf{FCLK}) \cite{Szeliski2006_fclk_extended} and inverse compositional (\textbf{ICLK}) \cite{Baker01ict} - that have been shown \cite{Baker01ict,Baker04lucasKanade_paper} to be equivalent to first order terms. Here, however, we show experimental results  (Sec. \ref{res_sm}) proving that they perform differently in practice. 
A more recent approach of this type is the Efficient Second order Minimization (\textbf{ESM}) \cite{Benhimane07_esm_journal} technique that uses gradients from both $ I_0 $ and $ I_t $ to make the best of ICLK and FCLK. 
Several extensions have also been proposed to these SMs to handle specific challenges like motion blur \cite{Park2009_esm_blur}, resolution degradation \cite{Ito2011_res_degrade}, better Taylor series approximation \cite{Keller2008} and optimal subset selection \cite{Benhimane2007}. Though these can be fit within our framework, either as distinct SMs or as variants thereof, they are not considered here for lack of space.

There are three main stochastic SMs in literature - approximate nearest neighbor (\textbf{NN}) \cite{Gu2010_ann_classifier_tracker,Dick13nn}, particle filters (\textbf{PF}) \cite{Isard98condensation,Kwon2009_affine_ois,Li2012_pf_affine_self_tuning_journal,Firouzi2014227_online_am,Kwon2014_sl3_aff_pf} and random sample consensus (\textbf{RANSAC}) \cite{Brown2007_ransac_stitching,Zhang2015_rklt}. Though less prone to getting stuck in local maxima than GD, their performance depends largely on the number and quality of random samples used and tends to be rather jittery and unstable due to the limited coverage of the search space. One approach to obtain better precision is to combine them with GD methods \cite{Dick13nn,Zhang2015_rklt} where results from the former are used as starting points for the latter. Three such combinations have been tested here as examples of hybrid or composite SMs - NN+ICLK (\textbf{NNIC}), PF+FCLK (\textbf{PFFC}) and RANSAC+FCLK (\textbf{RKLT}).



\section{Structural Similarity}
\parskip 0pt
\label{sec_ssim}
SSIM was originally introduced \cite{Wang04_ssim_original} to assess the loss in image quality incurred by compression methods like JPEG. It has been very popular in this domain since it closely mirrors the approach adopted by the human visual system to subjectively evaluate the quality of an image. 
SSIM between $ \mathbf{I_0} $ and $\mathbf{I_t}$  is defined as a product of 3 components:

{\footnotesize
\begin{align}
\label{eq_ssim_complete}
f_{ssim}
=
\left(
\dfrac{2\mu_t\mu_0+C_1}{\mu_t ^2+\mu_0^2+C_1}\right)^\alpha
\left(\dfrac{2\sigma_{t}\sigma_{0}+C_2}{\sigma_t ^2 + \sigma_0^2 + C_2}\right)^\beta
\left(\dfrac{\sigma_{t0}+C_3}{\sigma_{t}\sigma_{0}+C_3}\right)^\gamma
\end{align}
} %
where
$\mu_t$ is the mean and $ \sigma_{t}$ is the sample standard deviation of $ \mathbf{I_t} $ while
$ \sigma_{t0} $ is the sample \textit{covariance} between $ \mathbf{I_t} $ and  $ \mathbf{I_0} $. The three components of $ f_{ssim} $ from left to right are respectively used for luminance, contrast and structure comparison between the two patches. The positive constants $ \alpha,\beta,\gamma$ are used to assign relative weights to these components while $ C_1 $, $ C_2 $, $ C_3 $ are added to ensure their numerical stability with small denominators. Here, as in most practical implementations \cite{Wang04_ssim_original,Loza2009_ssim_pf,Loza2009a_ssim_hybrid,Loza11_ssim_lk_tracking,Wang2013_ssim_bc}, it is assumed that $ \alpha = \beta = \gamma = 1 $ and $ C_3 = \dfrac{C_2}{2} $ so that Eq. \ref{eq_ssim_complete} simplifies to:
\begin{align}
\label{eq_ssim}
f_{ssim}
=
\dfrac{
\left(2\mu_t\mu_0+C_1\right)
\left(2\sigma_{t0}+C_2\right)
}
{
\left(\mu_t ^2+\mu_0^2+C_1\right)
\left(\sigma_t ^2 + \sigma_0^2 + C_2\right)
}
\end{align}
\subsection{Newton's Method with SSIM}
\parskip 0pt
\label{newton_ssim}
Using SSIM with NN and PF is straightforward since these only need $ f_{ssim} $ to be computed between candidate patches. However, GD SMs also require its derivatives as they solve Eq \ref{reg_tracking} by estimating an incremental update $ \Delta\mathbf{p_t} $ to $ \mathbf{p_{t-1}} $ using some variant of Newton's method as:
\begin{align}
\label{eq_nwt_method}
\Delta\mathbf{p}_t=-\hat{\mathbf{H}}^{-1}\hat{\mathbf{J}}^T
\end{align}
where  $ \hat{\mathbf{J}} $ and $ \hat{\mathbf{H}} $ respectively are estimates for Jacobian  $  \mathbf{J} = \partial f/\partial \mathbf{p}$ and Hessian $  \mathbf{H} = \partial^2 f/\partial \mathbf{p}^2$ of $ f $ w.r.t. $ \mathbf{p} $. 
These can be further decomposed using chain rule as:
\begin{gather}
\label{eq_jac_basic}
\mathbf{J}
=
\dfrac{\partial f}{\partial \mathbf{I}} 
\dfrac{\partial I}{\partial \mathbf{p}}  
= 
\dfrac{\partial f}{\partial \mathbf{I}} 
\nabla \mathbf{I}
\dfrac{\partial  \mathbf{w}}{\partial \mathbf{p}}\\
\label{eq_hess_basic}
\mathbf{H}
=
\dfrac{\partial \mathbf{I}}{\partial \mathbf{p}}^T
\dfrac{\partial^2 f}{\partial \mathbf{I}^2}
\dfrac{\partial \mathbf{I}}{\partial \mathbf{p}}
+
\dfrac{\partial f}{\partial \mathbf{I}}
\dfrac{\partial^2 \mathbf{I}}{\partial \mathbf{p}^2}
\end{gather}
Of the right hand side terms in Eqs. \ref{eq_jac_basic} and \ref{eq_hess_basic}, only $\partial f/\partial \mathbf{I}  $ and $\partial^2 f/\partial \mathbf{I}^2  $ depend on $ f $ so the relevant expressions for $ f_{ssim} $ are presented below - general formulations corresponding to $ \mathbf{J} $ and $ \mathbf{H} $ in Eqs. \ref{eq_ssim_grad_final} and \ref{eq_ssim_hess_final} respectively, followed by specializations for $ \hat{\mathbf{J}} $ and $ \hat{\mathbf{H}} $ used by specific SMs. Detailed derivations are presented in the supplementary.

{\footnotesize
\begin{align}
\label{eq_ssim_grad_final}
\dfrac{\partial f_{ssim}}{\partial \mathbf{I_t}}
=   
\mathbf{f'}
=
\dfrac{2}{cd}
\left[
\left(\dfrac{a\mathbf{\bar{I}_0}-cf\mathbf{\bar{I}_t}}{N-1} + \dfrac{\mu_0b - \mu_tfd}{N} \right)
\right]
\end{align}
\begin{align}
\label{eq_ssim_hess_final}
\dfrac{\partial^2 f_{ssim}}{\partial \mathbf{I_t}^2}
&=   
\dfrac{2}{cd}
\bigg[
\dfrac{1}{N}
S_N
\left(
\dfrac{4}{N-1}
\left(
\mu_0\mathbf{\bar{I}_0}- \mu_tf\mathbf{\bar{I}_t}
\right)
-
\dfrac{3\mu_td}{2}\mathbf{f'}
-
\dfrac{fd}{N}
\right)\notag\\
&-
\dfrac{c}{N-1}
\left(
\dfrac{3}{2}\mathbf{f'}^T\mathbf{\bar{I}_t}
+
f\mathbb{I}
\right)
\bigg]
\end{align}
} %
with $ \mathbf{\bar{I}_t} =  \mathbf{I_t}-\mu_t, a=2\mu_t\mu_0+C_1 $, $ b=2\sigma_{t0}+C_2 $, $ c=\mu_t ^2+\mu_0^2+C_1, d=\sigma_t ^2 + \sigma_0^2 + C_2, f=f_{ssim} $ and  $ S_n\left(\mathbf{K}\right) $ denoting an $ n{\times} k $ matrix formed by stacking the $ 1\times k $ vector $ \mathbf{K} $ into rows.
The form of $\partial f/\partial \mathbf{I}$ used by the four variants of LK is identical except that ICLK requires the differentiation to be done w.r.t. $ \mathbf{I_0} $ instead of $ \mathbf{I_t} $ - the expressions for this are trivial to derive since SSIM is symmetrical. 
ESM was originally \cite{Benhimane07_esm_journal} formulated as using the mean of $\nabla \mathbf{I_0}$ and $\nabla \mathbf{I_t}$ to compute $ \mathbf{J} $ but, as this formulation is only applicable to SSD, a generalized version \cite{Brooks10_esm_ic,Scandaroli2012_ncc_tracking} is considered here that uses the \textit{difference} between FCLK and ICLK Jacobians.

It is generally assumed \cite{Baker04lucasKanade_paper,Benhimane07_esm_journal} that the second term of Eq. \ref{eq_hess_basic} is too costly to compute and too small near convergence to matter and so is omitted to give the Gauss Newton Hessian:
\begin{align}
\label{eq_hess_basic_gn}
\mathbf{H}_{gn}
=
\dfrac{\partial \mathbf{I}}{\partial \mathbf{p}}^T
\dfrac{\partial^2 f}{\partial \mathbf{I}^2}
\dfrac{\partial \mathbf{I}}{\partial \mathbf{p}}
\end{align}
Though $ \mathbf{H}_{gn} $ works very well for SSD (and in fact even better than $ \mathbf{H}$ \cite{Baker04lucasKanade_paper,Dame10_mi_thesis}), it is well known \cite{Dame10_mi_thesis,Scandaroli2012_ncc_tracking} to \textit{not} perform well with other AMs like MI, CCRE and NCC and we can confirm that the latter is true for SSIM and SPSS too. For these AMs, an approximation to $ \mathbf{H} $ \textit{after convergence} has to be used instead by assuming perfect alignment between the patches or $\mathbf{I^c}=\mathbf{I^*} $ . This approximation is here referred to as the \textbf{Self Hessian} and, as this substitution can be made by setting either $ \mathbf{I^c} = \mathbf{I_0}$ or $ \mathbf{I^*} = \mathbf{I_t} $, we get two forms which are respectively deemed to be ICLK and FCLK Hessians:

{\footnotesize
\begin{gather}
\label{eq_hess_iclk}
\hat{\mathbf{H}}_{ic}
=
\mathbf{H}_{self}^\mathbf{*}
=
\dfrac{\partial \mathbf{\mathbf{I_0}}}{\partial \mathbf{p}}^T
\dfrac{\partial^2 f(\mathbf{I_0}, \mathbf{I_0})}{\partial \mathbf{I}^2}
\dfrac{\partial \mathbf{\mathbf{I_0}}}{\partial \mathbf{p}}
+
\dfrac{\partial f(\mathbf{I_0}, \mathbf{I_0})}{\partial \mathbf{I}}
\dfrac{\partial^2 \mathbf{\mathbf{I_0}}}{\partial \mathbf{p}^2}\\
\label{eq_hess_fclk}
\hat{\mathbf{H}}_{fc}
=
\mathbf{H}_{self}^\mathbf{c}
=
\dfrac{\partial \mathbf{\mathbf{I_t}}}{\partial \mathbf{p}}^T
\dfrac{\partial^2 f(\mathbf{I_t}, \mathbf{I_t})}{\partial \mathbf{I}^2}
\dfrac{\partial \mathbf{\mathbf{I_t}}}{\partial \mathbf{p}}
+
\dfrac{\partial f(\mathbf{I_t}, \mathbf{I_t})}{\partial \mathbf{I}}
\dfrac{\partial^2 \mathbf{\mathbf{I_t}}}{\partial \mathbf{p}^2}
\end{gather}
} %
It is interesting to note that $ \mathbf{H}_{gn} $ has the exact same form as $ \mathbf{H}_{self} $ for SSD (since $ \partial f_{ssd}(\mathbf{I_0}, \mathbf{I_0})/\partial \mathbf{I} =  \partial f_{ssd}(\mathbf{I_t}, \mathbf{I_t})/\partial \mathbf{I} = \mathbf{0}$) so it seems that interpreting Eq. \ref{eq_hess_basic_gn} as the first order approximation of Eq. \ref{eq_hess_basic} for SSD  as in \cite{Baker04lucasKanade_paper,Dowson08_mi_ict,Dame10_mi_thesis} is incorrect and it should instead be seen as a special case of $ \mathbf{H}_{self} $.
Setting $ \mathbf{I_0} = \mathbf{I_t} $ simplifies Eq. \ref{eq_ssim_hess_final}  to:
\begin{align}
\dfrac{\partial^2 f_{ssim}(\mathbf{I_t}, \mathbf{I_t})}{\partial \mathbf{I_t}^2}
=
\dfrac{-2}{\bar{c}\bar{d}}
\left[
\dfrac{\bar{d}}{N^2}
+
\dfrac{\bar{c}}{N-1}\mathbb{I}
\right]
\end{align}
with $ \bar{c}=2\mu_t^2 + C_1 $ and $ \bar{d}=2\sigma_t^2 + C_2 $. Finally, FALK and IALK use the same form of $  \partial^2 f/\partial \mathbf{I}^2 $ as FCLK while ESM uses the \textit{sum} of $ \hat{\mathbf{H}}_{fc} $ and $\hat{\mathbf{H}}_{ic}$.
\subsection{Simplifying SSIM with pixelwise operations}
\parskip 0pt
In the formulation described so far, SSIM has been computed over the \textit{entire} patch - i.e. $ \mu_t $, $ \sigma_t $ and $ \sigma_{t0} $ have been computed over all $ N $ pixels in $ \mathbf{I_t} $ and $ \mathbf{I_0} $. In its original form \cite{Wang04_ssim_original}, however, the expression in Eq. \ref{eq_ssim} was applied to several corresponding \textit{sub windows} within the two patches - for instance $ 8\times 8 $ sub windows that are moved pixel-by-pixel over the entire patch - and the \textit{mean} of all resultant scores taken as the overall similarity score. For tracking applications, such an approach is not only impracticable from speed perspective, it presents another issue for GD SMs - presence of insufficient texture in these small sub windows may cause Eq. \ref{eq_nwt_method} to become ill posed if $ \mathbf{J}  $ and $ \mathbf{H} $ are computed for each sub window and then averaged.

As a result, the formulation used here considers only one end of the spectrum of variation of size and number of sub windows - a single sub window of the same size as the patch. Now, if the \textit{other} end of the spectrum is considered - $ N $ sub windows of size $ 1\times 1 $ each - then a different AM is obtained that may provide some idea about the effect of window wise operations while also being much simpler and faster. The resultant model is termed as \textbf{Sum of Pixelwise Structural Similarity} or \textbf{SPSS}. When considered pixel wise, $ \sigma_t $ and $ \sigma_{t0} $ become null while $ \mu_t$ becomes equal to the pixel value itself so that Eq. \ref{eq_ssim} simplifies to:
\begin{align}
\label{eq_spss}
f_{spss}
=
\sum\limits_{i=1}^{N}
\dfrac{
2\mathbf{I_t}(\mathbf{x}_\mathbf{it})
\mathbf{I_0}(\mathbf{x}_\mathbf{i0}) + C_1
}
{
\mathbf{I_t}(\mathbf{x}_\mathbf{it})^2
+
\mathbf{I_0}(\mathbf{x}_\mathbf{i0})^2
+
C_1
}
\end{align}
Similar to SSD, contributions from different pixels to $ f_{spss} $ are independent of each other so that each entry of $ \partial f_{spss}/\partial \mathbf{I_t}$ has contribution only from the corresponding pixel. This also holds true for each entry of the principal diagonal of $\partial^2 f_{spss}/\partial \mathbf{I_t}^2 $ (which is a diagonal matrix). Denoting the contributions of the $ i^{th} $ pixel to $  f_{spss} $, $ \partial f_{spss}/\partial \mathbf{I_t}$ and $\partial^2 f_{spss}/\partial \mathbf{I_t}^2 $ respectively as $ f_i $, $ f'_i $ and $ f''_i $, we get:
\begin{align}
\label{eq_spss_grad}
f'_i
&=
\dfrac{
2
(\mathbf{I_0}(\mathbf{x}_\mathbf{i0})
-
\mathbf{I_t}(\mathbf{x}_\mathbf{it})
f_{i}
)
}
{
\mathbf{I_t}(\mathbf{x}_\mathbf{it})^2
+
\mathbf{I_0}(\mathbf{x}_\mathbf{i0})^2
+
C_1
}\\
\label{eq_spss_hess}
f''_i
&=
\dfrac{
-2(
f_{i} + 3\mathbf{I_t}(\mathbf{x}_\mathbf{it})f'_{i}
)
}
{
\mathbf{I_t}(\mathbf{x}_\mathbf{it})^2
+
\mathbf{I_0}(\mathbf{x}_\mathbf{i0})^2
+
C_1
}\\
\label{eq_spss_self_hess}
f''_i(\mathbf{I_t}, \mathbf{I_t})
&=
\dfrac{-2}
{
2\mathbf{I_t}(\mathbf{x}_\mathbf{it})^2 + C_1
}
\end{align}

\section{Results and Analysis}
\label{experiments}
\subsection{Datasets}
Following four publicly available datasets
have been used to analyze the trackers:
\begin{enumerate}
\item Tracking for Manipulation Tasks (\textbf{TMT}) dataset \cite{Roy2015_tmt} that contains videos of some common tasks performed at several speeds and under varying lighting conditions. It has 109 sequences with 70592 frames. 
\item  Visual Tracking Dataset provided by \textbf{UCSB} \cite{Gauglitz2011_ucsb} that has 96 short sequences with a total of 6889 frames. These are more challenging than TMT but also somewhat artificial as they were created to represent specific challenges rather than realistic tasks.
\item \textbf{LinTrack} dataset \cite{Zimmermann2009_lintrack} that has 3 long sequences with a total of 12477 frames. These are more realistic than those in UCSB but also more difficult to track.
\item A collection of 28 challenging planar tracking sequences from several significant works in literature \cite{Kwon2014_sl3_aff_pf,Silveira2007_esm_lighting,Silveira2009_ilm_rgb_vs,Silveira2010_esm_ilm_rgb,Silveira14_esm_ilm_omni,Crivellaro2014_dft,Nebehay2015_cmt}. We call this the \textbf{PAMI} dataset after \cite{Kwon2014_sl3_aff_pf} from where several of the sequences originate. There are 16511 frames in all.
\end{enumerate} 
All of these datasets except PAMI have full pose (8 DOF) ground truth data which makes them suitable for evaluating high precision trackers that are the subject of this study. For PAMI, this data was generated using a combination of very high precision tracking and manual refinement \cite{Roy2015_tmt}.

\subsection{Evaluation Measures}
\textbf{Alignment Error} ($E_{AL}$) \cite{Dick13nn} has been used as the metric to compare tracking result with the ground truth since it accounts for fine misalignments of pose better than other measures like center location error and Jaccard index.
A tracker's overall accuracy is measured through its \textbf{success rate} (SR) which is defined as the fraction of total frames where $E_{AL}$ is less than a threshold of $t_p$ pixels. Formally, 
$ SR = |S|/|F| $
where $ S = \{f^{i} \in F : E_{AL}^{i} < t_p  \}$, $F$ is the set of all frames and $E_{AL}^{i}$ is the error in the $i^{th}$ frame $f^{i}$. 

Since there are far too many sequences to present results for each, an overall summary of performance is reported instead by averaging the SR over all sequences in the four datasets. In addition, to better utilize frames that follow a tracker's first failure in any sequence, we initialize trackers at $ 10 $ different evenly spaced frames in each sequence. Therefore the SR plots represent accumulated tracking performance over a total of $ |F| = 589380  $ frames, out of which $ 106469 $ are unique. 

Finally, the SR is evaluated for several values of $ t_p $ ranging from 0 to 20 and the resulting SR vs. $ t_p $ plot is studied to get an overall idea of how precise and robust a tracker is. 
As an alternative measure for robustness, reinitialization tests \cite{Kristan2016_vot16} were also conducted where a tracker is reinitialized after skipping 5 frames every time its $  E_{AL} $ exceeds 20 and the number of such reinitialization is counted.
Due to space constraints, these results are presented in the supplementary (available on MTF website).
 
\subsection{Parameters Used}
All results were generated using a fixed sampling resolution of $ 50{\times}50 $ irrespective the tracked object's size. Input images were smoothed using a Gaussian filter with a $ 5{\times}5 $ kernel before being fed to the trackers. Iterative SMs were allowed to perform a maximum of $ 30 $ iterations per frame but only as long as the L2 norm of the change in bounding box corners in each iteration exceeded $ 0.0001 $. Implementation details of NN, PF and RANSAC were taken from \cite{Dick13nn}, \cite{Kwon2014_sl3_aff_pf} and \cite{Zhang2015_rklt} respectively.
NN was run with 1000 samples and PF with 500 particles.
For RANSAC and RKLT, a $ 10{\times} 10 $ grid of sub patches was used and each sub patch was tracked by a 2 DOF FCLK tracker with a sampling resolution of $ 25{\times} 25 $. As in \cite{Wang04_ssim_original}, SSIM parameters are computed as $ C_1=(K_1L)^2 $ and $ C_2=(K_2L)^2 $ with $ K_1 =0.01$, $ K_2 =0.03$ and $ L=255$.  Learning based trackers (Sec. \ref{res_ssm}) were run using default settings provided by their respective authors. Speed tests were performed on a 4 GHz Intel Core i7-4790K machine with 16 GB of RAM.
\begin{figure*}[!htbp]
\begin{subfigure}{\textwidth}
  \centering
  \includegraphics[width=\textwidth]{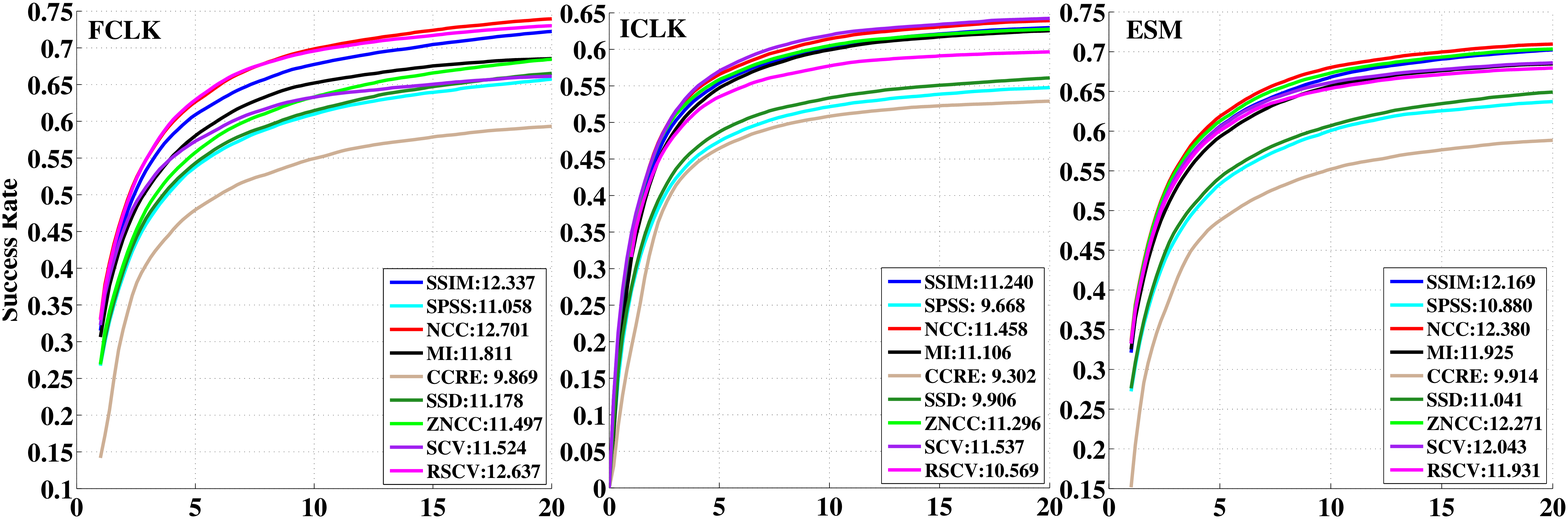}
\end{subfigure}
\begin{subfigure}{\textwidth}
  \centering
  \includegraphics[width=\textwidth]{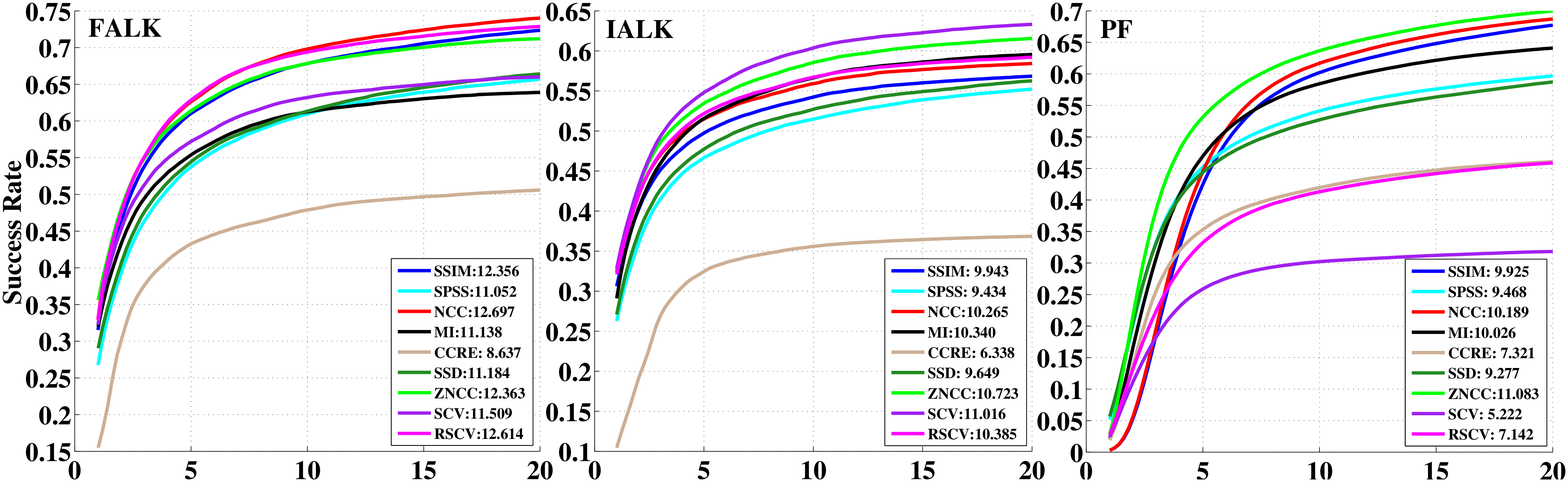}
\end{subfigure}
\begin{subfigure}{\textwidth}
  \centering
  \includegraphics[width=\textwidth]{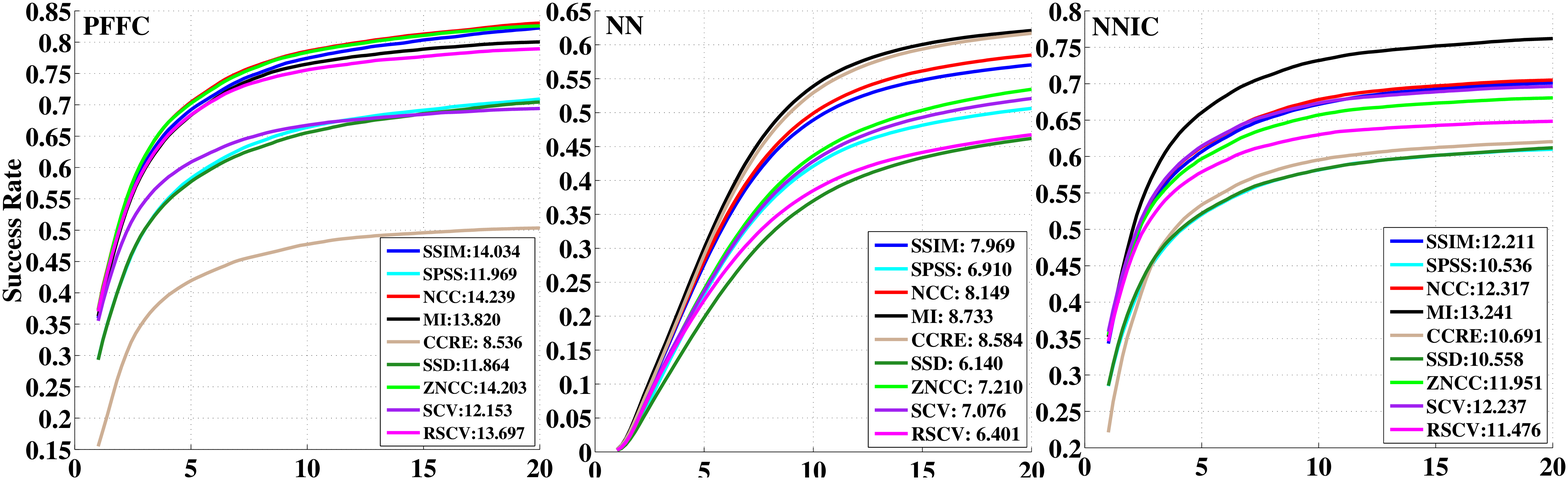}
\end{subfigure}
\begin{subfigure}{\textwidth}
  \centering
  \includegraphics[width=\textwidth]{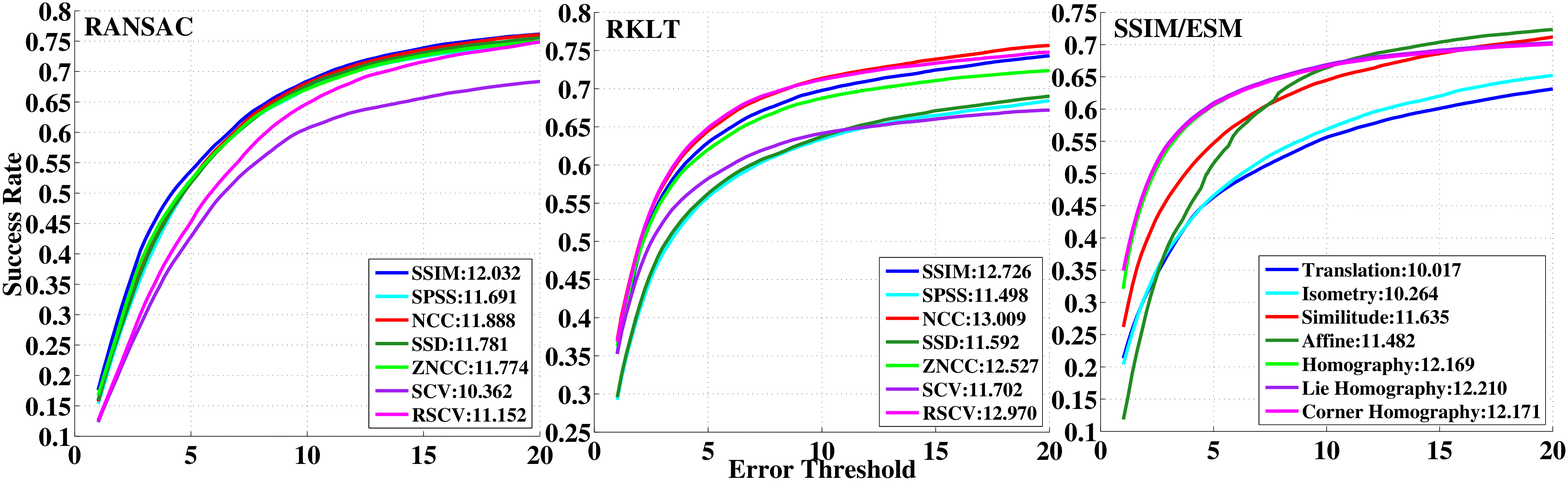}
\end{subfigure}
\caption{Success rates for AMs using different SMs with Homography as well as for different SSMs using SSIM with ESM (bottom right). Plot legends indicate areas under the respective curves. Best viewed on a high resolution screen.}
\label{fig_am}
\end{figure*}

\begin{figure}[t]
\includegraphics[width=0.5\textwidth]{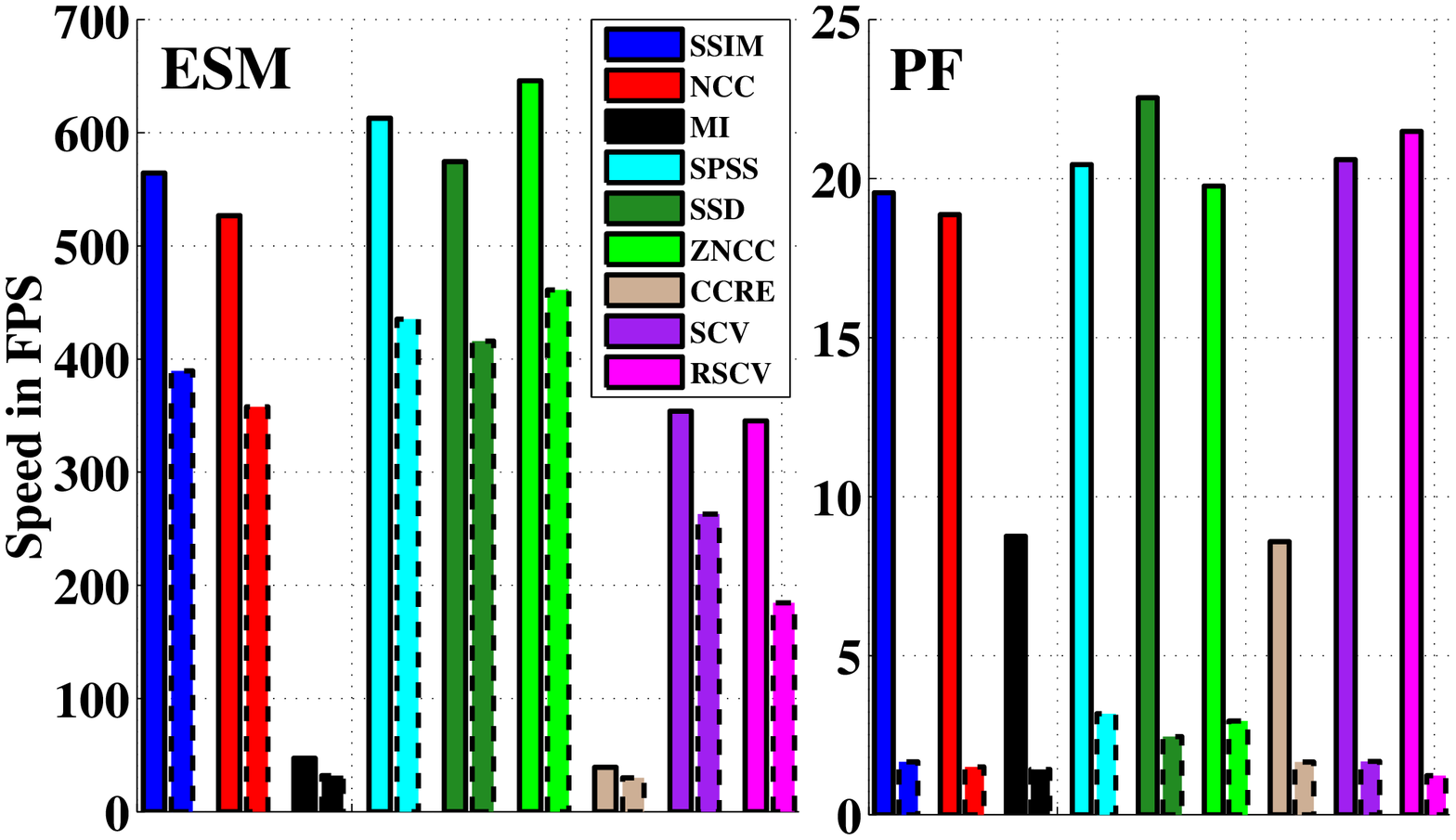}
\caption{Speeds of ESM and PF with homography. Solid and dotted lines show the means and standard deviations respectively.}
\label{fig_speed_esm_pf}
\end{figure}

\subsection{Results}
\label{results}
The results presented in this section are organized into three sections corresponding to the three sub modules. In each, we present and analyze results comparing different methods for the respective sub module.
\subsubsection{Appearance Models}
\label{res_am}
Fig. \ref{fig_am} shows the SR curves for all AMs using different SMs and homography SSM with plot legends also indicating the areas under the respective curves which are equivalent \cite{vCehovin2014_eval} to the average SR over this range of $ t_p $. Fig. \ref{fig_speed_esm_pf} shows the speeds of these AMs with ESM and PF as representatives of GD and stochastic SMs respectively.
LSCV is excluded to reduce clutter as it was found to perform very similarly to SCV. Its results are in the supplementary instead along with those for the ILMs. Further, RANSAC and RKLT do not include CCRE and MI results since both are far too slow for 100 of them to be run simultaneously in real time. Also, both perform \textit{very} poorly with these SMs, probably because the small sub patches used there \cite{Zhang2015_rklt} do not have sufficient information to be represented well by the joint histograms employed by these AMs. 

Several observations can be made here. Firstly, NCC is the best performer with all SMs except IALK and NN - IALK performs poorly with all robust AMs (Sec. \ref{res_sm}) while NN performs best with MI. Nevertheless, SSIM is usually equivalent to NCC or a close second showing that the latter is among the best AMs known. Also, as expected, SSIM is much better than SPSS with all SMs, with the latter only managing to match the performance of SSD on average. 
Further, though ZNCC is claimed to be equivalent to NCC \cite{Ruthotto2010_thes_ncc_equivalence} and also has a wider basin of convergence due to its SSD like formulation, it usually does \textit{not} perform as well as NCC. 

Secondly, in spite of being the most sophisticated and computationally expensive AM, CCRE is the worst performer with GD SMs and even MI is only slightly better than SSD on average. However, MI and CCRE are actually the best performing AMs with NN and MI is so with NNIC too. This shows that their poor performance with GD SMs is likely to be due to their narrower basin of convergence rather than an inherent weakness in the AMs themselves. This also explains MI's significant lead over CCRE with these SMs though the two differ only in the latter using a cumulative joint histogram. It seems likely that the additional complexity of CCRE along with the resultant invariance to appearance changes further \textit{reduces} its basin of convergence \cite{Dame10_mi_ict}. This discrepancy in performance between GD and stochastic SMs demonstrates the inadequacy of evaluating an AM with only one SM.

Thirdly, SCV outperforms RSCV with both inverse GD SMs - ICLK and IALK - though the reverse is true with both the forward ones - FCLK and FALK. This pattern also holds for stochastic and composite SMs - SCV is better with NN/NNIC where samples are generated from $ I_0 $ but worse with PF/PFFC where these come from $ I_t $. Also, their performance is very similar with ESM where information from both $ I_0 $ and $ I_t $ is used. This is probably because SCV performs likelihood substitution with $ \mathbf{I_0} $ \cite{Richa11_scv_original} while RSCV does so with $ \mathbf{I_t} $ \cite{Dick13nn}. Fourthly, the separation between AMs is narrower with RANSAC than other SMs as this SM depends more on the \textit{number} of sub patches used than on the tracker used for each. Conversely, PF shows maximum variation between AMs, thus indicating its strong reliance on $ f $.

Finally, SPSS is not much faster than SSIM with either SM though it has lower computational complexity. This is partly due to SSIM finding convergence in fewer iterations with GD SMs and partly due to the way Eigen \cite{eigenweb} optimizes matrix multiplications, many of which are used for computing $ f_{ssim} $ and its derivatives while those of $ f_{spss} $ have to be computed pixel by pixel. The same holds for SSD too.

\subsubsection{State Space Models}
\label{res_ssm}
\begin{figure*}[!htbp]
\begin{center}
\includegraphics[width=\textwidth]{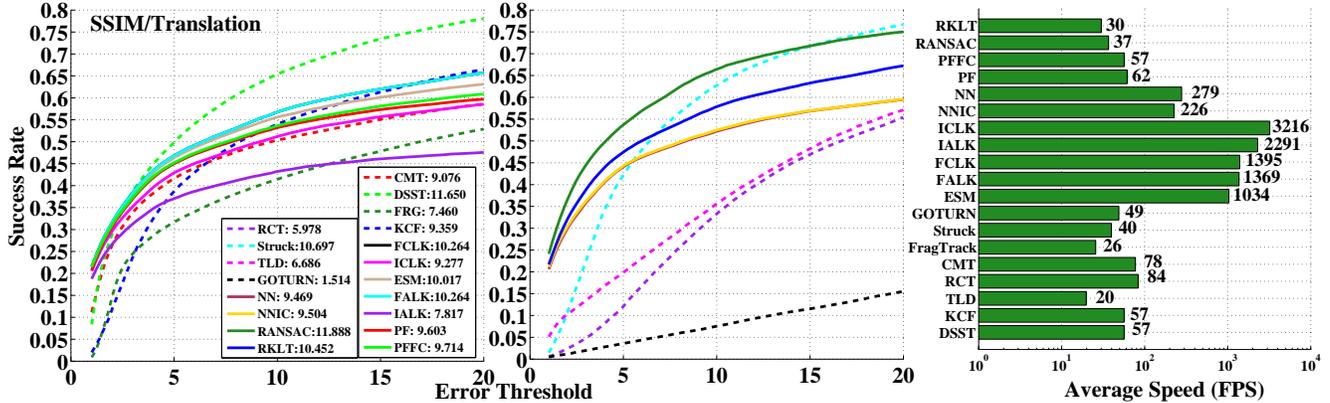}
\caption{Success Rates for SSIM using translation as well as for 8 OLTs. The former are shown with \textbf{solid} lines and the latter in \textbf{dashed} lines. The speed plot on the right has \textbf{logarithmic scaling} on the x axis for clarity though actual figures are also shown.}
\label{fig_lt_with_ssim_2r}
\end{center}
\end{figure*}
Results in this section follow a slightly different format from the other two due to the difference in the motivations for using low DOF SSMs - they are more robust due to reduced search space and faster due to the gradients of $ \mathbf{w} $ being less expensive to compute.
Limiting the DOF also makes them directly comparable to OLTs  which too have low DOF. As a result, 2 DOF RBTs were also tested with 8 state of the art OLTs \cite{Kristan2016_vot16} - DSST \cite{Danelljan2014_dsst}, KCF \cite{henriques2015high_kcf}, CMT \cite{Nebehay2015_cmt}, RCT \cite{Zhang2012_rct}, TLD \cite{Kalal12tld}, Struck \cite{hare2011struck}, FRG \cite{Adam2006_fragtrack} and GOTURN \cite{Held2016_goturn}. 
Lastly, in order to make the evaluations fair, \textit{lower DOF ground truth} has been used for all accuracy results in this section. This was generated for each SSM using least squares optimization to find the warp parameters that, when applied to the initial bounding box, will produces a warped box whose $E_{AL}$ with respect to the 8 DOF ground truth is as small as it is possible to achieve given the constraints of that SSM. 

Fig. \ref{fig_lt_with_ssim_2r} shows the results of these tests in terms of both accuracy and speed. 
As expected, all the OLTs have low SR for smaller $ t_p $ since they are less precise in general \cite{Kristan2016_vot16}. What is more interesting, however, is that none of them, with the exception of DSST and perhaps Struck, managed to surpass the best RBTs even for larger $ t_p $.
The superiority of DSST and Struck over other OLTs is consistent with results on VOT dataset \cite{Kristan2016_vot16}. However, the very poor performance of GOTURN \cite{Held2016_goturn}, which is one of the best trackers on that dataset, indicates a fundamental difference in the challenges involved in the two paradigms of tracking.
The speed plot shows another reason why OLTs are not suitable for high speed tracking applications - they are $ 10 $ to $ 30 $ times slower than RBTs except PF and RANSAC that are not implemented efficiently yet. 
It is not surprising that tracking based SLAM systems like SVO \cite{Forster2014_svo} use registration based trackers as they need to track hundreds of patches per frame.
To conclude the analysis in this section, we tested the performance of different SSMs against each other and the results are reported in the bottom right subplot of Fig. \ref{fig_am} using ESM with SSIM. Contrary to expectations, lower DOF SSMs are not better except perhaps affine and similitude that have slightly higher SR than homography for larger $ t_p $.
In fact, the SR curves of all low DOF SSMs approach those of homography as $ t_p $ increases which does indicate their higher robustness - though not as precise as homography, they  tend to be more resistant to complete failure.
Also, all three parameterizations of homography have almost identical performance, with their plots showing near perfect overlap. This suggests that the theoretical justification \cite{Benhimane07_esm_journal} for using ESM only with $\mathbb{SL}(3)$ has little practical significance.

\subsubsection{Search Methods}
\label{res_sm}
SR plots comparing different SMs for each AM are presented in the supplementary to save space as they contain the same data as Fig. \ref{fig_am}.
First fact to observe is that the four variants of LK do not perform identically. Though FCLK and FALK are indeed evenly matched, both ICLK and IALK are significantly worse, with ICLK being notably better than IALK. This is especially true for CCRE where it outperforms both additive SMs. This contradicts the equivalence between these variants that was reported in \cite{Baker04lucasKanade_paper} and justified there using both theoretical analysis and experimental results. The latter, however, were only performed on synthetic images and even the former used several approximations. Therefore, it is perhaps not surprising that this equivalence does not hold under real world conditions. 

Secondly, ESM fails to outperform FCLK for any AM except SCV and ZNCC and even there it does not lead by much. This fact too emerges in contradiction to the theoretical analysis in \cite{Benhimane07_esm_journal} where ESM was shown to have second order convergence and so should be better than first order methods like FCLK. 
Thirdly, both additive LK variants, and especially IALK, fare worse against compositional ones when using robust AMs compared to SSD-like AMs. This is probably because the Hessian after convergence approach used for extending Newton's method to these AMs does not make as much sense for additive formulations \cite{Dame10_mi_thesis}.
Lastly, PFFC is the best performing SM followed by NNIC and RKLT which proves the efficacy of composite SMs.


\section{Conclusions}
\label{conclusions}
We presented two new similarity measures for registration based tracking and also formulated a novel way to decompose trackers in this domain into three sub modules. We tested many different combinations of methods for each sub module to gain interesting insights into their strengths and weaknesses. Several surprising results were obtained that proved previously published theoretical analysis to be somewhat inaccurate in practice, thus demonstrating the usefulness of this approach. 
Finally, the open source tracking framework used for generating these results was made publicly available so these can be easily reproduced. 
{\small
\bibliographystyle{ieee}
\bibliography{E:/UofA/Thesis/References/references}

\begin{thebibliography}{10}\itemsep=-1pt

\bibitem{eigenweb}
{Eigen: A C++ template library for linear algebra}.
\newblock \url{http://eigen.tuxfamily.org}.
\newblock Accessed: 2017-01-10.

\bibitem{Adam2006_fragtrack}
A.~Adam, E.~Rivlin, and I.~Shimshoni.
\newblock {Robust fragments-based tracking using the integral histogram}.
\newblock In {\em 2006 IEEE Computer Society Conference on Computer Vision and
  Pattern Recognition (CVPR'06)}, volume~1, pages 798--805. IEEE, 2006.

\bibitem{Baker01ict}
S.~Baker and I.~Matthews.
\newblock {Equivalence and efficiency of image alignment algorithms}.
\newblock In {\em CVPR, IEEE Conference on}, volume~1, pages I--1090--I--1097
  vol.1, 2001.

\bibitem{Baker04lucasKanade_paper}
S.~Baker and I.~Matthews.
\newblock {Lucas-Kanade 20 Years On: A Unifying Framework}.
\newblock {\em IJCV}, 56(3):221--255, Feb 2004.

\bibitem{Bartoli2006}
A.~Bartoli.
\newblock {Direct image registration with gain and bias}.
\newblock {\em Contributions au recalage d’images et {\`a} la reconstruction
  3D de sc{\`e}nes rigides et d{\'e}formables}, page~4, 2006.

\bibitem{Bartoli2008}
A.~Bartoli.
\newblock {Groupwise geometric and photometric direct image registration}.
\newblock {\em PAMI, IEEE Transactions on}, 30(12):2098--2108, 2008.

\bibitem{Benhimane2007}
S.~Benhimane, A.~Ladikos, V.~Lepetit, and N.~Navab.
\newblock {Linear and quadratic subsets for template-based tracking}.
\newblock In {\em CVPR}, pages 1--6. IEEE, 2007.

\bibitem{Benhimane04esmTracking}
S.~Benhimane and E.~Malis.
\newblock {Real-time image-based tracking of planes using efficient
  second-order minimization}.
\newblock In {\em IEEE/RSJ International Conference on Intelligent Robots and
  Systems}, volume~1, pages 943--948 vol.1, Sept 2004.

\bibitem{Benhimane07_esm_journal}
S.~Benhimane and E.~Malis.
\newblock {Homography-based 2D Visual Tracking and Servoing}.
\newblock {\em Int. J. Rob. Res.}, 26(7):661--676, July 2007.

\bibitem{Bookstein1989_tps}
F.~L. Bookstein.
\newblock {Principal warps: Thin-plate splines and the decomposition of
  deformations}.
\newblock {\em PAMI, IEEE Transactions on}, (6):567--585, 1989.

\bibitem{Brooks10_esm_ic}
R.~Brooks and T.~Arbel.
\newblock {Generalizing Inverse Compositional and ESM Image Alignment}.
\newblock {\em IJCV}, 87(3):191--212, May 2010.

\bibitem{Brown2007_ransac_stitching}
M.~Brown and D.~G. Lowe.
\newblock {Automatic panoramic image stitching using invariant features}.
\newblock {\em International journal of computer vision}, 74(1):59--73, 2007.

\bibitem{vCehovin2014_eval}
L.~{\v{C}}ehovin, M.~Kristan, and A.~Leonardis.
\newblock {Is my new tracker really better than yours?}
\newblock In {\em IEEE Winter Conference on Applications of Computer Vision},
  pages 540--547. IEEE, 2014.

\bibitem{Cobzas2009_ic_3d_tracking}
D.~Cobzas, M.~Jagersand, and P.~Sturm.
\newblock {3D SSD tracking with estimated 3D planes}.
\newblock {\em Image and Vision Computing}, 27(1):69--79, 2009.

\bibitem{Crivellaro2014_dft}
A.~Crivellaro and V.~Lepetit.
\newblock {Robust 3d tracking with descriptor fields}.
\newblock In {\em 2014 IEEE Conference on Computer Vision and Pattern
  Recognition}, pages 3414--3421. IEEE, 2014.

\bibitem{Dame10_mi_thesis}
A.~Dame.
\newblock {\em {A unified direct approach for visual servoing and visual
  tracking using mutual information}}.
\newblock PhD thesis, University of Rennes, 2010.

\bibitem{Dame10_mi_ict}
A.~Dame and E.~Marchand.
\newblock {Accurate real-time tracking using mutual information}.
\newblock In {\em IEEE International Symposium on Mixed and Augmented Reality},
  pages 47--56, 2010.

\bibitem{Danelljan2014_dsst}
M.~Danelljan, G.~H{\"a}ger, F.~Khan, and M.~Felsberg.
\newblock {Accurate scale estimation for robust visual tracking}.
\newblock In {\em British Machine Vision Conference, Nottingham, September 1-5,
  2014}. BMVA Press, 2014.

\bibitem{Dick13nn}
T.~Dick, C.~Perez, M.~Jagersand, and A.~Shademan.
\newblock {Realtime Registration-Based Tracking via Approximate Nearest
  Neighbour Search}.
\newblock In {\em Proceedings of Robotics: Science and Systems}, Berlin,
  Germany, June 2013.

\bibitem{Dowson08_mi_ict}
N.~Dowson and R.~Bowden.
\newblock {Mutual Information for Lucas-Kanade Tracking (MILK): An Inverse
  Compositional Formulation}.
\newblock {\em PAMI}, 30(1):180--185, Jan 2008.

\bibitem{Firouzi2014227_online_am}
H.~Firouzi and H.~Najjaran.
\newblock {Adaptive on-line similarity measure for direct visual tracking}.
\newblock {\em Image and Vision Computing}, 32(4):227 -- 236, 2014.

\bibitem{Forster2014_svo}
C.~Forster, M.~Pizzoli, and D.~Scaramuzza.
\newblock {SVO: Fast semi-direct monocular visual odometry}.
\newblock In {\em Robotics and Automation (ICRA), 2014 IEEE International
  Conference on}, pages 15--22. IEEE, 2014.

\bibitem{Gauglitz2011_ucsb}
S.~Gauglitz, T.~H{\"o}llerer, and M.~Turk.
\newblock {Evaluation of interest point detectors and feature descriptors for
  visual tracking}.
\newblock {\em International journal of computer vision}, 94(3):335--360, 2011.

\bibitem{Gu2010_ann_classifier_tracker}
S.~Gu, Y.~Zheng, and C.~Tomasi.
\newblock {Efficient visual object tracking with online nearest neighbor
  classifier}.
\newblock In {\em Computer vision--ACCV 2010}, pages 271--282. Springer, 2010.

\bibitem{Hager98parametricModels}
G.~D. Hager and P.~N. Belhumeur.
\newblock {Efficient Region Tracking With Parametric Models of Geometry and
  Illumination}.
\newblock {\em IEEE Transactions on Pattern Analysis and Machine Intelligence},
  20(10):1025--1039, October 1998.

\bibitem{hare2011struck}
S.~Hare, A.~Saffari, and P.~H. Torr.
\newblock {Struck: Structured output tracking with kernels}.
\newblock In {\em 2011 International Conference on Computer Vision}, pages
  263--270. IEEE, 2011.

\bibitem{Hartley04MVGCV}
R.~Hartley and A.~Zisserman.
\newblock {\em {Multiple View Geometry in Computer Vision}}.
\newblock Cambridge University Press, second edition, March 2004.

\bibitem{Held2016_goturn}
D.~Held, S.~Thrun, and S.~Savarese.
\newblock {\em {Learning to Track at 100 FPS with Deep Regression Networks}},
  pages 749--765.
\newblock Springer International Publishing, Cham, 2016.

\bibitem{henriques2015high_kcf}
J.~F. Henriques, R.~Caseiro, P.~Martins, and J.~Batista.
\newblock {High-speed tracking with kernelized correlation filters}.
\newblock {\em IEEE Transactions on Pattern Analysis and Machine Intelligence},
  37(3):583--596, 2015.

\bibitem{Isard98condensation}
M.~Isard and A.~Blake.
\newblock {CONDENSATION - conditional density propagation for visual tracking}.
\newblock {\em International Journal of Computer Vision}, 29:5--28, 1998.

\bibitem{Ito2011_res_degrade}
E.~Ito, T.~Okatani, and K.~Deguchi.
\newblock {Accurate and robust planar tracking based on a model of image
  sampling and reconstruction process}.
\newblock In {\em Mixed and Augmented Reality (ISMAR), 2011 10th IEEE
  International Symposium on}, pages 1--8. IEEE, 2011.

\bibitem{Jain2005_ncc_z_score}
A.~Jain, K.~Nandakumar, and A.~Ross.
\newblock {Score normalization in multimodal biometric systems}.
\newblock {\em Pattern recognition}, 38(12):2270--2285, 2005.

\bibitem{Jepson2003_online_am}
A.~D. Jepson, D.~J. Fleet, and T.~F. El-Maraghi.
\newblock {Robust online appearance models for visual tracking}.
\newblock {\em IEEE Transactions on Pattern Analysis and Machine Intelligence},
  25(10):1296--1311, 2003.

\bibitem{Kalal12tld}
Z.~Kalal, K.~Mikolajczyk, and J.~Matas.
\newblock {Tracking-Learning-Detection}.
\newblock {\em IEEE Transactions on Pattern Analysis and Machine Intelligence},
  34(7):1409--1422, \#jul\# 2012.

\bibitem{Keller2008}
Y.~Keller and A.~Averbuch.
\newblock {Global parametric image alignment via high-order approximation}.
\newblock {\em Computer Vision and Image Understanding}, 109(3):244--259, 2008.

\bibitem{Kristan2016_vot16}
M.~Kristan, A.~Leonardis, J.~Matas, M.~Felsberg, et~al.
\newblock {\em {The Visual Object Tracking VOT2016 Challenge Results}}, pages
  777--823.
\newblock Springer International Publishing, Cham, 2016.

\bibitem{Kwon2014_sl3_aff_pf}
J.~Kwon, H.~S. Lee, F.~C. Park, and K.~M. Lee.
\newblock {A geometric particle filter for template-based visual tracking}.
\newblock {\em Pattern Analysis and Machine Intelligence (PAMI), IEEE
  Transactions on}, 36(4):625--643, 2014.

\bibitem{Kwon2009_affine_ois}
J.~Kwon, K.~M. Lee, and F.~C. Park.
\newblock {Visual tracking via geometric particle filtering on the affine group
  with optimal importance functions}.
\newblock In {\em Computer Vision and Pattern Recognition, 2009. CVPR 2009.
  IEEE Conference on}, pages 991--998. IEEE, 2009.

\bibitem{Li2012_pf_affine_self_tuning_journal}
M.~Li, T.~Tan, W.~Chen, and K.~Huang.
\newblock {Efficient object tracking by incremental self-tuning particle
  filtering on the affine group}.
\newblock {\em IEEE Transactions on Image Processing}, 21(3):1298--1313, 2012.

\bibitem{Liu14_olt_survey}
Q.~Liu, X.~Zhao, and Z.~Hou.
\newblock {Survey of single-target visual tracking methods based on online
  learning}.
\newblock {\em IET Computer Vision}, 8(5):419--428, October 2014.

\bibitem{Loza2009_ssim_pf}
A.~{\L}oza, L.~Mihaylova, D.~Bull, and N.~Canagarajah.
\newblock {Structural similarity-based object tracking in multimodality
  surveillance videos}.
\newblock {\em Machine Vision and Applications}, 20(2):71--83, 2009.

\bibitem{Loza2009a_ssim_hybrid}
A.~Loza, F.~Wang, M.~A. Patricio, J.~Garc{\'\i}a, and J.~M. Molina.
\newblock {Optimised Particle Filter Approaches to Object Tracking in Video
  Sequences}.
\newblock In {\em Methods and Models in Artificial and Natural Computation},
  pages 486--495. Springer, 2009.

\bibitem{Loza11_ssim_lk_tracking}
A.~Loza, F.~Wang, J.~Yang, and L.~Mihaylova.
\newblock {Video object tracking with differential Structural SIMilarity
  index}.
\newblock In {\em ICASSP, IEEE International Conference on}, pages 1405--1408,
  May 2011.

\bibitem{Lucas81lucasKanade}
B.~D. Lucas and T.~Kanade.
\newblock {An Iterative Image Registration Technique with an Application to
  Stereo Vision}.
\newblock In {\em 7th International Joint Conference on Artificial
  intelligence}, volume~2, pages 674--679, 1981.

\bibitem{Nebehay2015_cmt}
G.~Nebehay and R.~Pflugfelder.
\newblock {Clustering of Static-Adaptive correspondences for deformable object
  tracking}.
\newblock In {\em Proceedings of the IEEE Conference on Computer Vision and
  Pattern Recognition}, pages 2784--2791, 2015.

\bibitem{Park2009_esm_blur}
Y.~Park, V.~Lepetit, and W.~Woo.
\newblock {ESM-Blur: Handling \& rendering blur in 3D tracking and
  augmentation}.
\newblock In {\em Proceedings of the 2009 8th IEEE International Symposium on
  Mixed and Augmented Reality}, pages 163--166. IEEE Computer Society, 2009.

\bibitem{Richa14_scv_constraints}
R.~Richa, M.~Souza, G.~Scandaroli, E.~Comunello, and A.~von Wangenheim.
\newblock {Direct visual tracking under extreme illumination variations using
  the sum of conditional variance}.
\newblock In {\em Image Processing (ICIP), 2014 IEEE International Conference
  on}, pages 373--377, Oct 2014.

\bibitem{Richa11_scv_original}
R.~Richa, R.~Sznitman, R.~Taylor, and G.~Hager.
\newblock {Visual tracking using the sum of conditional variance}.
\newblock In {\em IROS, IEEE/RSJ International Conference on}, pages
  2953--2958, Sept 2011.

\bibitem{Richa12_robust_similarity_measures}
G.~H. Rog´erio~Richa, Raphael~Sznitman.
\newblock {Robust Similarity Measures for Gradient-based Direct Visual
  Tracking}.
\newblock Technical report, CIRL, June 2012.

\bibitem{Ross08ivt}
D.~A. Ross, J.~Lim, R.-S. Lin, and M.-H. Yang.
\newblock {Incremental Learning for Robust Visual Tracking}.
\newblock {\em IJCV}, 77(1-3):125--141, May 2008.

\bibitem{Roy2015_tmt}
A.~Roy, X.~Zhang, N.~Wolleb, C.~Perez~Quintero, and M.~Jagersand.
\newblock {Tracking benchmark and evaluation for manipulation tasks}.
\newblock In {\em Robotics and Automation (ICRA), 2015 IEEE International
  Conference on}, pages 2448--2453. IEEE, 2015.

\bibitem{Ruthotto2010_thes_ncc_equivalence}
L.~Ruthotto.
\newblock {Mass-preserving registration of medical images}.
\newblock {\em German Diploma Thesis (Mathematics), Institute for Computational
  and Applied Mathematics, University of M{\"u}nster}, 2010.

\bibitem{Scandaroli2012_ncc_tracking}
G.~G. Scandaroli, M.~Meilland, and R.~Richa.
\newblock {Improving NCC-based Direct Visual Tracking}.
\newblock In {\em ECCV}, pages 442--455. Springer, 2012.

\bibitem{Shashua2001_q_warp}
A.~Shashua and Y.~Wexler.
\newblock {Q-warping: Direct computation of quadratic reference surfaces}.
\newblock {\em IEEE Transactions on Pattern Analysis and Machine Intelligence},
  23(8):920--925, 2001.

\bibitem{Shum00_fc}
H.-Y. Shum and R.~Szeliski.
\newblock {Construction of Panoramic Image Mosaics with Global and Local
  Alignment}.
\newblock {\em IJCV}, 36(2):101--130.

\bibitem{Silveira14_esm_ilm_omni}
G.~Silveira.
\newblock {Photogeometric Direct Visual Tracking for Central Omnidirectional
  Cameras}.
\newblock {\em Journal of Mathematical Imaging and Vision}, 48(72), October
  2014.

\bibitem{Silveira2007_esm_lighting}
G.~Silveira and E.~Malis.
\newblock {Real-time visual tracking under arbitrary illumination changes}.
\newblock In {\em CVPR. IEEE Conference on}, pages 1--6, 2007.

\bibitem{Silveira2009_ilm_rgb_vs}
G.~Silveira and E.~Malis.
\newblock {Visual servoing from robust direct color image registration}.
\newblock In {\em 2009 IEEE/RSJ International Conference on Intelligent Robots
  and Systems}, pages 5450--5455. IEEE, 2009.

\bibitem{Silveira10_esm_ext2}
G.~Silveira and E.~Malis.
\newblock {Unified Direct Visual Tracking of Rigid and Deformable Surfaces
  Under Generic Illumination Changes in Grayscale and Color Images}.
\newblock {\em International Journal of Computer Vision}, 89(1):84--105, 2010.

\bibitem{Silveira2010_esm_ilm_rgb}
G.~Silveira and E.~Malis.
\newblock {Unified direct visual tracking of rigid and deformable surfaces
  under generic illumination changes in grayscale and color images}.
\newblock {\em International journal of computer vision}, 89(1):84--105, 2010.

\bibitem{singh16_mtf}
A.~Singh and M.~Jagersand.
\newblock {Modular Tracking Framework: A Unified Approach to Registration based
  Tracking}.
\newblock arXiv:1602.09130 [cs.CV], 2016.

\bibitem{singh16_modular_results}
A.~Singh, A.~Roy, X.~Zhang, and M.~Jagersand.
\newblock {Modular Decomposition and Analysis of Registration based Trackers}.
\newblock In {\em CRV}, June 2016.

\bibitem{Smeulders2014_tracking_survey}
A.~W. Smeulders, D.~M. Chu, R.~Cucchiara, S.~Calderara, A.~Dehghan, and
  M.~Shah.
\newblock {Visual tracking: An experimental survey}.
\newblock {\em Pattern Analysis and Machine Intelligence, IEEE Transactions
  on}, 36(7):1442--1468, 2014.

\bibitem{Szeliski2006_fclk_extended}
R.~Szeliski.
\newblock {Image alignment and stitching: A tutorial}.
\newblock {\em Foundations and Trends{\textregistered} in Computer Graphics and
  Vision}, 2(1):1--104, 2006.

\bibitem{Szeliski1997_spline}
R.~Szeliski and J.~Coughlan.
\newblock {Spline-based image registration}.
\newblock {\em IJCV}, 22(3):199--218, 1997.

\bibitem{Trummer2008_gklt}
M.~Trummer, J.~Denzler, and C.~Munkelt.
\newblock {Guided KLT Tracking Using Camera Parameters in Consideration of
  Uncertainty}.
\newblock In {\em International Conference on Computer Vision and Computer
  Graphics}, pages 252--261. Springer, 2008.

\bibitem{Wang2007_ccre_registration}
F.~Wang and B.~C. Vemuri.
\newblock {Non-rigid multi-modal image registration using cross-cumulative
  residual entropy}.
\newblock {\em IJCV}, 74(2):201--215, 2007.

\bibitem{Wang2013_ssim_bc}
X.~Wang, C.~Ning, A.~Shi, and G.~Lv.
\newblock {An improved similarity measure in particle filters for robust object
  tracking}.
\newblock In {\em Image and Signal Processing (CISP), 2013 6th International
  Congress on}, volume~1, pages 46--50. IEEE, 2013.

\bibitem{Wang04_ssim_original}
Z.~Wang, A.~Bovik, H.~Sheikh, and E.~Simoncelli.
\newblock {Image quality assessment: from error visibility to structural
  similarity}.
\newblock {\em Image Processing, IEEE Transactions on}, 13(4):600--612, April
  2004.

\bibitem{Zhang2012_rct}
K.~Zhang, L.~Zhang, and M.-H. Yang.
\newblock {Real-time compressive tracking}.
\newblock In {\em European Conference on Computer Vision}, pages 864--877.
  Springer, 2012.

\bibitem{Zhang2015_rklt}
X.~Zhang, A.~Singh, and M.~Jagersand.
\newblock {RKLT: 8 DOF real-time robust video tracking combing coarse RANSAC
  features and accurate fast template registration}.
\newblock In {\em CRV}, pages 70--77. IEEE, 2015.

\bibitem{Zimmermann2009_lintrack}
K.~Zimmermann, J.~Matas, and T.~Svoboda.
\newblock {Tracking by an optimal sequence of linear predictors}.
\newblock {\em Pattern Analysis and Machine Intelligence, IEEE Transactions
  on}, 31(4):677--692, 2009.

\end{thebibliography}
}
\end{document}